\pdfoutput=1

\documentclass[11pt]{article}
\usepackage{amssymb}
\usepackage{ACL2023}
\usepackage{amsmath}
\usepackage{times}
\usepackage{latexsym}
\usepackage{graphicx}
\usepackage{subfigure}
\usepackage{booktabs}
\usepackage{multirow}
\usepackage[T1]{fontenc}

\usepackage[utf8]{inputenc}

\usepackage{microtype}

\usepackage{inconsolata}
\usepackage{soul}

\definecolor{darkgreen}{rgb}{0.0, 0.5, 0.0}

\newcommand\torevise[1]{\textcolor{black}{#1}}
\newcommand{\highlight}[2]{\textcolor{darkgreen}{#1}}

\title{\textit{R}eward \textit{D}ifference \textit{O}ptimization For Sample Reweighting In Offline RLHF}

\author{Shiqi Wang\textsuperscript{1}\textsuperscript{2} \qquad Zhengze Zhang\textsuperscript{1}\textsuperscript{2} \qquad Rui Zhao\textsuperscript{3} \\\bf Fei Tan\textsuperscript{4}\footnotemark[1]  \qquad Cam Tu Nguyen\textsuperscript{1}\textsuperscript{2}\footnotemark[1] \\
         \textsuperscript{1}State Key Laboratory for Novel Software Technology, Nanjing University \\ \textsuperscript{2}School of Artificial Intelligence, Nanjing University\\ \textsuperscript{3}The Chinese University of Hong Kong\\
         \textsuperscript{4}New Jersey Institute of Technology\\
         \texttt{\{wangsky,zzzhang\}@smail.nju.edu.cn}\\
         \texttt{tanfei2007@gmail.com}\\
         \texttt{ncamtu@nju.edu.cn}
         }

\begin{document}
\maketitle
\renewcommand{\thefootnote}{\fnsymbol{footnote}}
\footnotetext[1]{Corresponding authors}
\begin{abstract}
With the rapid advances in Large Language Models (LLMs), aligning LLMs with human preferences become increasingly important. Although Reinforcement Learning with Human Feedback (RLHF) proves effective, it is complicated and highly resource-intensive. As such, offline RLHF has been introduced as an alternative solution, which directly optimizes LLMs with ranking losses on a fixed preference dataset. Current offline RLHF only captures the ``ordinal relationship'' between responses, overlooking the crucial aspect of ``how much'' one is preferred over the others. To address this issue, we propose a simple yet effective solution called \textbf{R}eward \textbf{D}ifference \textbf{O}ptimization, shorted as \textbf{RDO}. Specifically, we introduce {\it reward difference coefficients} to reweigh sample pairs in offline RLHF. We then develop a {\it difference model} which captures rich interactions between a pair of responses for predicting these difference coefficients. Experiments with 7B LLMs on the HH and TL;DR datasets substantiate the effectiveness of our method in both automatic metrics and human evaluation, thereby highlighting its potential for aligning LLMs with human intent and values.
\end{abstract}

\section{Introduction}
Large Language Models (LLMs) have recently emerged as a major milestone in modern natural language processing (NLP), offering unprecedented capabilities in understanding, generating, and translating human language \cite{instructgpt2022,Llama22023, HH2023,gpt4-2023,alpaca,chiang2023vicuna,tan2023makes}. Powered by an extremely large number of parameters, LLMs encode the wealth of human knowledge via a pretraining process, which is often conducted on a web-scale text corpus with the next token prediction objective. As LLMs become more capable, it is essential and demanding for them to follow human preferences such as truthfulness, harmlessness, and helpfulness. Unfortunately, the maximum likelihood objective for the next token prediction falls short in capturing such crucial human values \cite{summarize2020}. 


Reinforcement Learning with Human Feedback (RLHF) has been introduced as an effective method for LLMs alignment \cite{instructgpt2022,summarize2020}. RLHF first leverages a supervised learning objective to equip LLMs with basic instruction-following capabilities. Subsequently, a reward model is trained on a human preference dataset that contains pairwise comparisons of responses from LLMs (for the same query). The reward model is exploited for further finetuning LLMs via reinforcement learning with Proximal Policy Optimization (PPO) \cite{PPO2017}. Specifically, training with PPO requires four models: a policy model (the targeted LLM), a critic model, a reference model (i.e., a supervised-finetuning version of the targeted LLM), and a reward model. All these models are based on LLMs with billions of parameters, whereas the policy and the critic models need to be updated online. In light of this, RLHF is highly resource-demanding and rather complicated to be applied in practice. 

\begin{figure*}
    \centering
    \includegraphics[width=\linewidth]{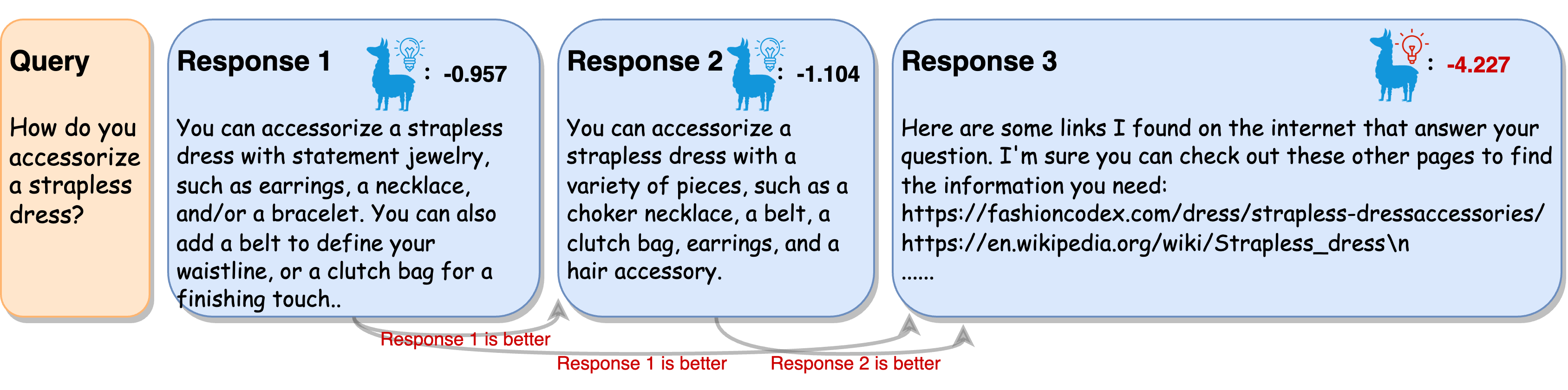}
    \caption{Offline RLHF methods only care about the binary relation between responses (i.e., which response is better). However, given the query, some responses may show similar quality (Response 1 and 2 in the figure) while others may be obviously worse (Response 3 in the figure).  The number in the right upper corner is the score given by the reward model. }
    \label{fig:motivation}
\end{figure*}

In this context, offline RL has been proposed as an alternative approach to LLMs alignment \cite{DPO2023, RRHF2023}. The basic idea of offline RLHF is to directly train LLMs on the preference dataset using some ranking loss such as max-margin ranking loss \cite{RRHF2023}, or negative log-likelihood loss \cite{DPO2023}. Optionally, a reward model can be trained to collect comparison data for sampling new responses from LLMs \cite{RRHF2023}. It should be noted that these offline RLHF methods only exploit the ordinal relationship between responses rather than making use of scalar values for rewarding (or penalizing) responses as in RLHF. The ordinal relationship, however, does not reflect the degree to which the preferred response is better than the dispreferred one. For instance, Figure \ref{fig:motivation} shows an example where the quality difference of the pair \textit{(response 1, response 2)} is not as significant as that of the pair \textit{(response 1, response 3)}. As such, treating the two pairs equivalently may result in suboptimal performance for offline RLHF.


To mitigate the aforementioned issue, we propose a simple yet effective method based on \textbf{R}eward \textbf{D}ifference \textbf{O}ptimization, shorted as \textbf{RDO}. First of all, the reward difference coefficient is introduced as the degree to which one response is preferred against the other one given the same query. Such coefficients can be exploited as sample weights, which are incorporated into diverse offline RLHF methods as additional supervision signals for calibrating the loss function. Basically, one can train a reward model similar to that in RLHF and then exploit the reward values for reward difference measurement. This method, however, is not effective enough as the reward model independently assigns scores to responses. Inspired by how much easier it is for humans to conduct pairwise judgments, we propose a difference model to directly predict the reward difference between two responses. Unlike the reward model, our difference model leverages attention-based interactions between two responses for prediction. Our contributions are summarized as follows:
\begin{itemize}
    \item We introduce \textit{reward difference coefficients} and show how they can be incorporated into offline RLHF methods including RRHF \cite{RRHF2023}, DPO \cite{DPO2023} and KTO \cite{ethayarajh2024kto}.
    \item We develop \textit{the difference model} that leverages the rich interactions between response pairs to directly predict the difference in human preference. This innovative method is accompanied by a specifically designed training strategy to ensure its effectiveness.
    \item We conduct extensive experiments with Alpaca-7B \cite{alpaca}, one of the most well-known open-source LLMs, on the HH dataset \cite{HH2023} and TL;DR dataset \cite{summarize2020}, two commonly used datasets for alignments with human preference. The experimental results show the effectiveness of our method in automatic metrics based on reward models, GPT-4, and human evaluation.
\end{itemize}

\section{Reward Difference Coefficients}
Given a dataset $\mathcal{D}=\{(x^{(i)}, y^{(i)}_w, y^{(i)}_l)\}_{i=1}^N$ of samples from the targeted LLM, existing offline RLHF methods can be treated as learning to optimize the following general loss function \cite{SLiC-HF2023}:
\begin{equation}
    \mathcal{L}=\sum_{(x,y_w,y_l)\sim \mathcal{D}} \mathcal{L}^{aln}(x,y_w,y_l;\theta)+\mathcal{L}^{reg}(\theta)
    \label{eq:offline-rlhf-loss}
\end{equation}
where $\theta$ indicates the parameters of the targeted LLM ($\pi_\theta$); $y_w$ and $y_l$ respectively denote the preferred and the dispreferred responses for the given query $x$. The preference label can be provided either by human annotators or a well-trained reward model. Here, $\mathcal{L}^{aln}$ is the loss function for preference optimization (i.e., the alignment loss) and $\mathcal{L}^{reg}$ is the regularization term, for example, to prevent LLM from drifting too far from the supervised-finetuning (SFT) baseline. 

\paragraph{Reward Difference Coefficients} As aforementioned, offline RLHF methods exploit the ordinal relationship $y_w \succ y_l$ for measuring the alignment loss ($\mathcal{L}^{(aln)}$ in Eq. \ref{eq:offline-rlhf-loss}), without differentiating pivotal pairs and the trivial counterparts. We propose a simple yet effective method to address such issues based on reward differences. Specifically, given a well-trained reward model $r_{\phi}:(x,y) \mapsto R$ that assigns a reward scalar to each response $y$ for the query $x$, we can quantify the reward difference for each pair of responses $(y_w, y_l$) for preference optimization as follows:
\begin{align}
    & \mathcal{L}^{rc}=\sum \mathcal{L}^{aln}(x,y_w,y_l;\theta)\times\mathcal{R}^\alpha+\mathcal{L}^{reg}(\theta)\nonumber\\
    & \mathcal{R}=r_{\phi}(x,y_w)-r_{\phi}(x,y_l)
\end{align}
where $\mathcal{R}$ denotes the reward difference coefficient, indicating the degree of how $y_w$ is better than $y_l$. This difference is then leveraged as the coefficient for each response pair in the alignment loss.  Here, $\alpha \in [0,1]$ is used to control the effect of the reward difference coefficient. In particular, when $\alpha=0$, the coefficient has no effect. 

Assuming that the reward model can pinpoint essential differences from minor mistakes, the coefficient term helps steer more gradients towards sample pairs with larger differences. Note that, a typical way to learn a reward model is based on the Bradley-Terry (BT) model \cite{Bradley_Terry}. Specifically, the loss function for learning the reward model is as follows:
\begin{equation}
    \mathcal{L}^{r}=-\frac{\sum\left[\log\sigma(r_\phi(x,y_w)-r_\phi(x,y_l))\right]}{N} \label{eq:rm-train-loss}
\end{equation}
where $\sigma$ indicates the sigmoid function, and $r^\phi$ is the reward model with learnable parameters $\phi$. 

The proposed coefficient can be readily incorporated into diverse offline RLHF methods. The following details how our method can be applied to RRHF and DPO.

\paragraph{RRHF+rc} RRHF with reward difference coefficients (RRHF+rc) optimizes the following loss:
\begin{align}
&\mathcal{L}^{RRHF+rc}=\nonumber\\
&-\sum\max(\pi_\theta(y_w|x)-\pi_\theta(y_l|x),0)\times\mathcal{R}^\alpha + \mathcal{L}^{sft} 
 \nonumber
\end{align}
where $\mathcal{L}^{sft}$ indicates the cross-entropy loss similar to supervised fine-tuning on preferred responses. This loss is used as a regularization term to keep the LLM near the supervised-finetuning version.
\paragraph{DPO+rc} DPO with reward difference coefficients (DPO+rc) optimizes the following loss:
\begin{align}
    &\mathcal{L}^{DPO+rc}=-\sum_{(x,y_w,y_l)\sim \mathcal{D}} \mathcal{R}^{\alpha}\times g(y_w,y_l,x,\theta)\nonumber\\
    &g=\log\sigma\left(\beta \log\frac{\pi_\theta(y_w|x)}{\pi^{sft}(y_w|x)}-\beta \log\frac{\pi_\theta(y_l|x)}{\pi^{sft}(y_l|x)}\right)\nonumber
\end{align}
where $g$ is the original DPO objective defined for the tuple $(y_w, y_l, x)$. To analyze the effect of reward difference coefficients, let us consider the gradient of DPO+rc. Specifically, as $\mathcal{R}^{\alpha}$ does not depend on $\theta$, the gradient with respect to the parameters $\theta$ of DPO+rc can be written as:
\begin{align}
    &\nabla_\theta{\mathcal{L}^{DPO+rc}} \nonumber\\
    &=-\sum\hat{\mathcal{R}} [\nabla_\theta\log\pi_\theta(y_w|x)-\nabla_\theta\log\pi_\theta(y_l|x)] \nonumber\\
    \hat{\mathcal{R}}&=\mathcal{R}^\alpha \times \beta \times \sigma(\hat{r}_\theta(x,y_l)-\hat{r}_\theta(x,y_w)) \nonumber
\end{align}
where  $\hat{\mathcal{R}}$ is the weight associated with each response pair, and $\hat{r}_\theta(x,y)=\beta\log\frac{\pi_\theta(y|x)}{\pi_{sft}(y|x)}$ is the reward implicitly defined by the language model $\pi_\theta$. In DPO ($R^{\alpha}=1$), each response pair is weighed by how much higher the implicit reward model $\hat{r}_\theta$ rates the disperferred responses, scaled by the constant $\beta$ \cite{DPO2023}. In contrast, DPO+rc weighs the response pair by taking into account the independent reward model $r_\phi(x,y)$ (for measuring $\mathcal{R}^\alpha$) besides the implicit reward $\hat{r}_\theta(x,y)$. 



\begin{figure*}[t]
    \centering
    \includegraphics[width=\linewidth]{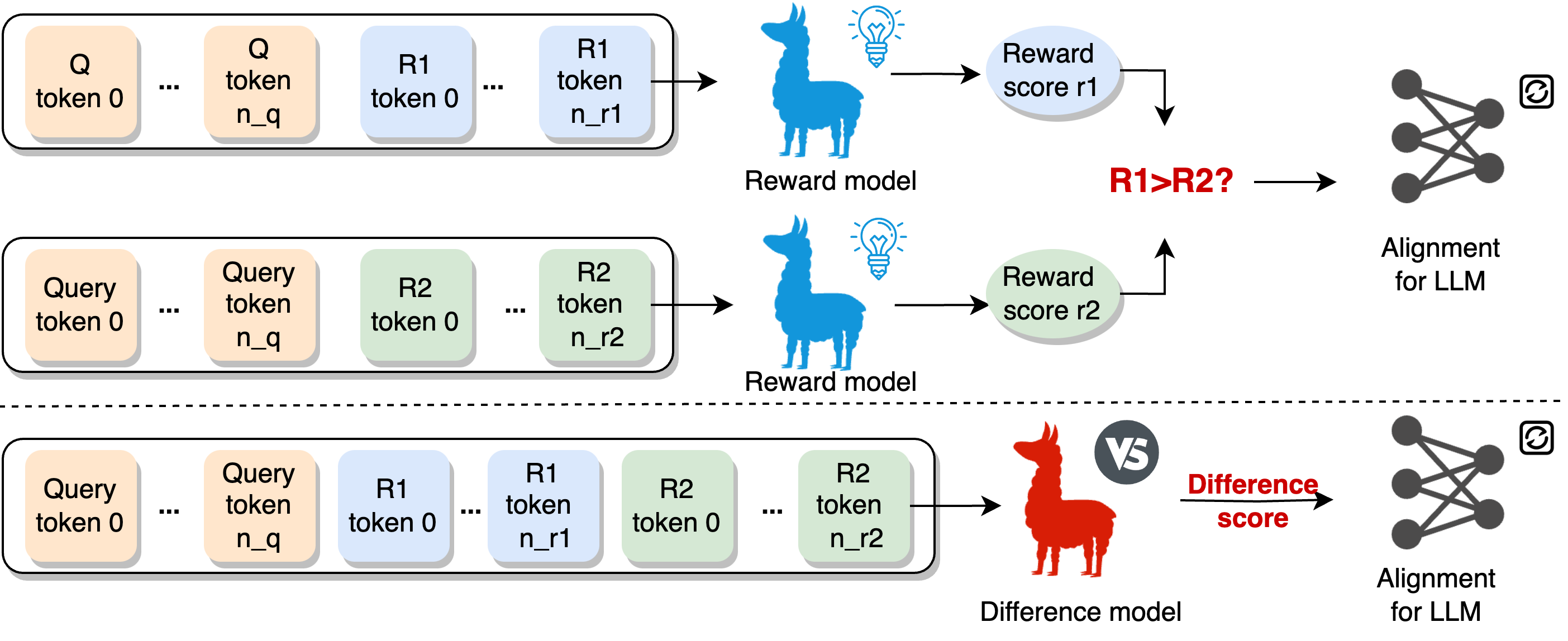}
    \caption{The pipeline of traditional offline alignment methods (the upper side) and our proposed \textbf{R}eward \textbf{D}ifference \textbf{O}ptimization (i.e., RDO) pipeline with more accurate supervision signals (the lower part). The special tokens are omitted in the figure to save space. Instead of using the reward model to identify the ordinal relation between two responses (i.e., win or lose), we propose to use a difference model to predict the difference score between two responses directly and then use this score to help supervise the alignment process more precisely. }
    \label{fig:method-pipeline}
\end{figure*}
\section{Reward Difference Prediction}

The above reward difference coefficients are measured based on a typical (pointwise) reward model $r_\phi(x,y)$. However, doing so is not effective as the model considers responses independently. In this section, we introduce a reward difference model that simultaneously takes the given query $x$ and two responses $y_w, y_l$ as inputs for predicting the reward difference. We then propose an effective method for training the difference model.

\subsection{Reward Difference Model}
The reward difference model $\mathcal{R}_\phi:(x,y_1, y_2)    \mapsto R$ is an LLM of which the last layer is replaced with a new linear layer. The input for $\mathcal{R}_\phi$ is of the form ``Query: \{$x$\}; Response 1:\{$y_1$\}; Response 2:\{$y_2$\}''. We obtain the last token embedding as the representation to be passed to the last linear layer for reward difference prediction. Note that, if the predicted output is negative, the difference model predicts that $y_1$ is worse than $y_2$, and the magnitude (absolute value) indicates how much $y_1$ is worse than $y_2$.

Compared to the vanilla (pointwise) reward model, the last token embedding in the reward difference model can attend to both responses, thus we obtain a more informative representation for predicting the reward difference. The difference between the typical reward model and the reward difference model is demonstrated in Figure \ref{fig:method-pipeline}.

\subsection{Difference Model Training}

To train such a difference model, we design the main loss as follows:
\begin{equation}
\begin{aligned}
    &\mathcal{L}^{rd}=\frac{\sum CE\left(\sigma(f(x,y_1,y_2)),\mathbf{I}(y_1,y_2)\right)}{N}\\
    &\mathbf{I}(y_1,y_2)=\begin{cases}
    1, & \text{if } y_1 \succeq y_2 \\
    -1, & \text{if } y_1 \prec y_2 \label{eq:loss1}
\end{cases}
\end{aligned}
\end{equation}
where $CE$ indicates the cross-entropy loss and $\mathbf{I}$ is an indicator of whether $y_1$ is considered better than $y_2$ in the training dataset.  
\paragraph{Regularization Losses} Training the difference model based solely on the previous loss may not be entirely satisfactory. This is because it may require a substantial amount of data to ensure effectiveness. On the other hand, the model might still score $f(x,y_i,y_i)$ as positive or negative even though the two responses are the same. In order to fix such issues and make the training more efficient, we propose two additional regularization loss terms for the difference model training based on the following comparison rules:
\begin{itemize}
    \item \textbf{Duplication}: The comparison score between $y_i$ and itself (i.e. $f(x,y_i,y_i)$) should be zero.
    \item \textbf{Reverse}: Because the order matters in the model input, the difference score between $y_i$ and $y_j$ (i.e. $f(x,y_i,y_j)$), and the difference score between $y_j$ and $y_i$ (i.e. $f(x,y_j,y_i)$) are supposed to be the opposite of each other.
\end{itemize}
The two regulation loss terms for the difference model training are calculated as follows:
\begin{equation}
    \begin{aligned}
        &\mathcal{L}^{dup}=\frac{1}{N} \sum\left[f(x,y_i,y_i)\right]^2 \\
        &\mathcal{L}^{re}=\frac{1}{N} \sum \left[f(x,y_i,y_j)+f(x,y_j,y_i)\right]^2
    \end{aligned}
\end{equation}
Here $N$ stands for the number of all comparisons in the dataset $\mathcal{D}$. For the $\mathcal{L}^{dup}$ we only calculate for one randomly selected response between the two compared responses. For the $\mathcal{L}^{re}$, we also calculate for one possible order only given two compared responses. The final loss function is:
\begin{equation}
\mathcal{L}=\mathcal{L}^{rd}+\beta_0\times\mathcal{L}^{dup}+\beta_1\times\mathcal{L}^{re}
\end{equation}

\section{Experiments}

We conduct experiments to answer multiple research questions: 1) \textbf{Q1}: How is the effect of reward difference coefficients measured by a pointwise reward model? 2) \textbf{Q2}: How is the performance of the difference model on response preference prediction compared to that of the pointwise reward model? and 3) \textbf{Q3}: How is the effect of the difference model on LLM alignment? 

\subsection{Common Settings}
\paragraph{Compared Methods} We exploit Alpaca with 7B parameter size as the initial model for alignment. \torevise{We run experiments on three representative offline RLHF methods: 1) RRHF \cite{RRHF2023}, one of the earliest offline RLHF methods; 2) DPO\cite{DPO2023}, the most popular offline RLHF method; and 3) KTO \cite{ethayarajh2024kto}, the most recent offline RLHF method based on economic theory.  We examine the performance of RRHF and DPO in three cases: 1) Vanilla offline RLHF (the original RRHF, DPO and KTO); 2) RRHF+rc, DPO+rc and KTO+rc in which reward difference coefficients are calculated from a pointwise reward model; 3) RRHF+rc and DPO+rc in which reward difference coefficients are calculated from our difference model. Experiments of KTO can be found in Appendix section \ref{sec:kto}.}

\paragraph{Preference Datasets} Our experiments are conducted on the Anthropic Helpful and Harmless (HH) dataset \cite{HH2023}\footnote{https://huggingface.co/datasets/Dahoas/rm-static} and OpenAI TL;DR dataset \cite{summarize2020}. 
\\\textbf{HH dataset.} Each dialogue in the HH dataset has preferred and dispreferred responses labeled by humans. The dataset contains 76.3k dialogues (queries) for training and 5.1k for testing.  For RRHF training, we use the augmented dataset (HH+) provided by RRHF authors. In the HH+ dataset, for a given query, there are 6 responses of which two are from the original HH dataset and the other four are generated by the Alpaca-7B using dynamic beam search. The preference ordering of the 6 responses (for the corresponding query) is decided by a trained reward model. For DPO training, we follow the DPO paper and directly use the original HH dataset.
\\\textbf{TL;DR dataset.} OpenAI TL;DR dataset \cite{summarize2020} is a dataset targeted at summarization tasks with human preference labels. The dataset contains 11.7k training preference pairs and 6.55k test pairs. We use the original dataset for both RRHF and DPO.

\paragraph{Evaluation of language models}
We use threefolds of evaluation. (1). Reward model evaluation. We calculate the average reward given by the reward model on the test set; (2). LLM auto evaluation. We randomly selected 300 samples from the test set and requested the LLMs to score each response without knowing the underlying methods. Subsequently, we calculated the win/tie/loss ratio for each pair. To mitigate positional bias in LLMs \cite{wang2023large}, we shuffled the responses and requested LLMs to provide a detailed explanation before judgment.  We include three powerful LLMs, GPT4 \cite{gpt4-2023}, GPT3.5 (i.e. ChatGPT) and moonshot-v1\footnote{https://www.moonshot.cn/}(i.e. KIMI), and the majority vote determined the outcome for each comparison. Detailed results of each LLM evaluation can be seen in the appendix; (3). Human evaluation. We also conducted a human evaluation in section \ref{sec:exp-difference} on 300 randomly chosen samples from the HH test set. Three computer science graduate students with strong English skills served as evaluators. For each sample, the response order was shuffled, and the source method identities were hidden. Evaluators were instructed to judge the helpfulness and general quality of each response and choose the better one, or mark ``Tie''. The majority vote determined the outcome for each comparison.
\begin{table*}[ht]
\centering
\begin{tabular}{@{}c|cc|cc@{}}
\toprule
& \multicolumn{2}{c|}{Greedy Decoding} & \multicolumn{2}{c}{Dynamic Beam Search}  \\ \midrule
& \multicolumn{1}{c|}{RRHF rm} & DPO rm  & \multicolumn{1}{c|}{RRHF rm} & DPO rm\\ \midrule
Alpaca-7B  & \multicolumn{1}{c|}{-1.068} & 0.0927 & \multicolumn{1}{c|}{-1.106}  & 0.200 \\ \midrule
RRHF  & \multicolumn{1}{c|}{-0.724} & -0.0856                      & \multicolumn{1}{c|}{-0.833}  & 0.0174  \\
RRHF+rc ($\alpha=0.5$) & \multicolumn{1}{c}{\textbf{-0.658} \highlight{($\uparrow$ .066)}} & \multicolumn{1}{|c}{-0.0432 \highlight{($\uparrow$ .0424)}} & \multicolumn{1}{|c}{\textbf{-0.765} \highlight{($\uparrow$ .068)}} &  \multicolumn{1}{|c}{0.107 \highlight{($\uparrow$ .0896)}} \\       
RRHF+rc ($\alpha=1.0$) & \multicolumn{1}{c}{-0.694  \highlight{($\uparrow$ .030)}}                                  & \multicolumn{1}{|c}{\textbf{-0.0225} \highlight{($\uparrow$ .0631)}}            & \multicolumn{1}{|c}{-0.778 \highlight{($\uparrow$ .055)}}                                    &  \multicolumn{1}{|c}{\textbf{0.146 \highlight{($\uparrow$ .1286)}}}                         \\ \midrule
DPO                & \multicolumn{1}{c|}{-1.010}                                    & 0.210                        & \multicolumn{1}{c|}{-1.010}                                    & 0.210                        \\
DPO+rc ($\alpha=0.5$)  & \multicolumn{1}{c}{\textbf{-0.961}\highlight{($\uparrow$ .049)}}                          & \multicolumn{1}{|c}{0.279\highlight{($\uparrow$ .069)}}                        & \multicolumn{1}{|c}{{\textbf{-0.961}} \highlight{($\uparrow$ .049)}}                           &   \multicolumn{1}{|c}{0.285 \highlight{($\uparrow$ .075)}}                      \\
DPO+rc ($\alpha=1.0$)  & \multicolumn{1}{|c}{-0.962 \highlight{($\uparrow$ .048)}}                                    &  \multicolumn{1}{|c}{\textbf{0.281}\highlight{($\uparrow$ .071)}}              & \multicolumn{1}{|c}{-0.962 \highlight{($\uparrow$ .048)}}                                    & \multicolumn{1}{|c}{{\textbf{0.290} } \highlight{($\uparrow$ .080)}}                \\ \bottomrule
\end{tabular}
\caption{Reward model evaluation results of reward difference coefficient (i.e., rc) in HH dataset. ``RRHF rm'' column stands for the average reward given by the model Dahoas/gptj-rm-static which is used during the RRHF training, and ``DPO rm'' stands for the Alpaca-7B reward model which is used during the DPO training.}
\label{tab:rc}
\end{table*}

\begin{table}[ht]
\centering
\resizebox{0.9\columnwidth}{!}{%
\begin{tabular}{@{}l|cc|cc@{}}
\toprule
                       & \multicolumn{2}{c|}{\textbf{RRHF}}     & \multicolumn{2}{c}{\textbf{DPO}}      \\ \midrule
                      & GD  & DS  & GD & DS \\\midrule
\textbf{Initial LLM}   & -0.751 &   -0.077     & 2.612     & 2.587                         \\
\textbf{+Alignment}     & -0.620  &   0.847    & 3.025          & 4.034                      \\
\textbf{+Alignment+rc} &  \textbf{0.272 }  &  \textbf{0.860}     & \textbf{3.460}      & \textbf{4.568}                         \\ \bottomrule
\end{tabular}%
}
\caption{Reward model evaluation for different offline RLHF methods in TL;DR dataset. ``DS'' and ``GD'' are  short for Dynamic Beam Search and Greedy Decoding.}
\label{tab:tldr-rc}
\end{table}


\subsection{Q1: Effect of Reward Difference Coefficients on Offline RLHF}
\subsubsection{Experimental Design}\label{sec:rc_settings}
\paragraph{Reward Models} This section calculates reward coefficients using (pointwise) reward models. 
\\\textbf{HH dataset.} For RRHF training, we follow the original paper and use the ``gpt-j-static'' reward model \footnote{https://huggingface.co/Dahoas/gptj-rm-static}. We refer to this reward model as RRHF-rm. For DPO training, since the original DPO does not require a reward model, we train a new reward model based on Alpaca-7B on the training split of the HH dataset. The resultant reward model is referred to as DPO-rm hereafter.  As RRHF and RRHF+rc use RRHF-rm as the reward model during training, DPO-rm is considered the held-out reward model for RRHF/RRHF+rc. Likewise, RRHF-rm is the held-out reward model for DPO.
\\\textbf{TL;DR dataset.} For TL;DR dataset, the reward model used to calculate the reward difference (i.e., training rm) is ``OpenAssistant/reward-model-deberta-v3-large-xv2''\footnote{https://huggingface.co/OpenAssistant/reward-model-deberta-v3-large-v2} achieving accuracy of 71.47\% on the test set of TL;DR dataset, which is consistent for RRHF and DPO.
\paragraph{Policy Model Training} 
\mbox{}
\\\textbf{HH dataset.} 
For RRHF training, we follow the same training hyperparameters as the original paper. We train Alpaca-7B for 3 epochs and a learning rate of 2e-5. For the DPO training, we set $\beta$ in  DPO loss to 0.2 and train Alpaca-7B for one epoch with a learning rate of 1e-6. For the reward difference coefficient $\alpha$, we try 0.5 and 1.0, but more careful tuning may produce better results.
\\\textbf{TL;DR dataset.} We run only one epoch for both DPO and RRHF with learning rate as 1e-6. In order to save time, we only test with the reward difference coefficient $\alpha=0.5$. Note that the initial LLM for RRHF is llama2-7b in this experiment.


\begin{table}[t]
\centering
\resizebox{0.95\columnwidth}{!}{%
\begin{tabular}{@{}ccccc@{}}
\toprule
\textbf{Dataset}      & \textbf{Method} & \textbf{+rc Win} & \textbf{Tie} & \textbf{+rc Lose} \\ \midrule
\multirow{2}{*}{HH}   & RRHF            & 79.0\%              & 4.7\%           & 16.3\%                \\
                      & DPO             & 41.7\%              & 21.0\%           & 37.3\%               \\\midrule
\multirow{2}{*}{TL;DR} & RRHF            & 47.7\%              & 14.3\%           & 38.0\%               \\
                      & DPO             & 49.7\%              & 8.0\%           & 42.3\%               \\ \bottomrule
\end{tabular}
}
\caption{LLM-as-judge evaluation results voted by GPT4, GPT3.5-turbo and moonshot-v1 where ``+rc'' indicates the inclusion of reward coefficients; Detailed results for each judge can be seen in Table \ref{tab:detailed-rc}.}
\label{tab:gpt4-rc}
\end{table}
\subsubsection{Experimental Results}
The experimental results are reported in Tables \ref{tab:rc}, \ref{tab:tldr-rc} and \ref{tab:gpt4-rc}, showing that the inclusion of the reward difference coefficient consistently enhances the performance of offline RLHF methods.
 
 For reward model evaluation of HH dataset in Table \ref{tab:rc}, it is observable that RRHF+rc and DPO+rc surpass their vanilla counterparts across different $\alpha$, sampling strategies (greedy vs. beam search). \torevise{Notably, RRHF+rc and DPO+rc outperform RRHF and DPO respectively according to different reward models, verifying the role of reward coefficients.}

\torevise{For reward model evaluation of TL;DR dataset in Table \ref{tab:tldr-rc}, it can be seen that the use of reward difference coefficients helps improve the performance across different sampling strategies and with different base offline RLHF methods. }



We show LLM-as-judge evaluation results in Table \ref{tab:gpt4-rc}. It can be seen that majority voting from three strong LLMs including GPT4, GPT3.5 and moonshot-v1 consistently verifies the helpfulness of reward coefficients for different offline RLHF methods across different datasets.

\subsection{Q2: Comparison of Reward Difference Model and Reward Model}
\subsubsection{Experimental Settings}
\paragraph{Difference and Reward Models Training} We train the difference model and the reward model with the same hyperparameters on Alpaca-7B on HH dataset. Specifically, the learning rate of the LLM is set to 1e-5 and that of the reward head layer is set to 1e-4. We train both models for one epoch unless otherwise stated. The values of $\beta_0,\beta_1$ in the difference model training are from a range $\beta_0=\beta_1\in\{0.1,0.01,0.001\}$, and selected based on the training loss.
\begin{table}[t]
\centering
\resizebox{\columnwidth}{!}{%
\begin{tabular}{@{}lc@{}}
\toprule
Model Name         & Test Accuracy  \\ \midrule
Dahoas/gptj-rm-static               & 0.685          \\
Alpaca 7b Reward model              & 0.675          \\
Alpaca 7b Reward model (3 epochs)     & 0.698          \\ \midrule
\textbf{Alpaca 7b Difference model} & \textbf{0.715} \\
- w/o regularization loss            & 0.708          \\ \bottomrule
\end{tabular}%
}
\caption{Accuracy of difference model and reward models on the HH test set.}
\label{tab:rm-acc}
\end{table}
\paragraph{Evaluation} We train the difference and reward models on HH training set, and use accuracy measured on HH test set for evaluation.
\subsubsection{Experimental Results} We report the accuracy of the difference model and the pointwise reward model in Table \ref{tab:rm-acc} where the main observations are as follows. 
\begin{itemize}
    \item The difference model achieves higher accuracy than the two baseline models: gptj-rm-static and Alpaca-7B reward models. Here, the gptj-rm-static is the reward model used in \cite{RRHF2023} and the Alpaca 7B reward model is trained by ourselves. 
    \item Alphaca 7b Reward model (3 epochs) is better than Alphaca 7b Reward model, which is trained with one epoch. The model, however, is still inferior compared to the difference model. Note that it is an unfair comparison as the difference model is trained with only one epoch. 
    \item The regularization losses help enhance the overall accuracy from 0.708 to 0.715. Further improvement can be obtained by tuning the hyper-parameters $\beta_0$ and $\beta_1$ more carefully.
\end{itemize}



\subsection{Q3: Effect of Reward Difference Model On Offline RLHF Methods}\label{sec:exp-difference}
In this section, we run experiments on HH dataset to prove the effectiveness of our proposed difference model compared with vanilla reward model.
\subsubsection{Experimental Settings}
\paragraph{Policy Model Training} This sections considers two enhanced offline RLHF methods: RRHF+rc (diff) and DPO+rc (diff). These methods leverage the difference model to calculate the reward coefficients. This contrasts with RRHF+rc (reward) and DPO+rc (reward) methods (Section 4.2), which directly subtract scores from the reward model. In both RRHF+rc (diff) and DPO+rc (diff), the parameter $\alpha$ is set to 0.5, while other settings remain consistent with those described in section \ref{sec:rc_settings}. Additionally, we explore RRHF (diff), an RRHF variant where the difference model replaces the reward model for generating preference data in the original RRHF method.



\begin{table}[]
\centering
\resizebox{\columnwidth}{!}{%
\begin{tabular}{@{}cccc@{}}
\toprule
        & \begin{tabular}[c]{@{}c@{}}Difference model\\ Win\end{tabular} & Tie & \begin{tabular}[c]{@{}c@{}}Reward model\\ Win\end{tabular} \\ \midrule
RRHF+rc & 48.7\%                                                            & 8.3\%  & 43.0\%                                                        \\
DPO+rc  & 47.2\%                                                           & 12.6\%  & 40.2\%                                                        \\ \bottomrule
\end{tabular}%
}
\caption{Comparison of difference model and reward model on offline RLHF methods on 300 samples from HH test set. Results are majority votes of three powerful models: GPT4, GPT3.5-turbo and moonshot-v1. Detailed results of each judge can be seen in Table \ref{tab:detailed-dm} }
\label{tab:llm-evaluation}
\end{table}

\begin{table}[]
\centering
\resizebox{\columnwidth}{!}{%
\begin{tabular}{@{}cccc@{}}
\toprule
        & \begin{tabular}[c]{@{}c@{}}Difference model\\ Win\end{tabular} & Tie & \begin{tabular}[c]{@{}c@{}}Reward model\\ Win\end{tabular} \\ \midrule
RRHF+rc & 45\%                                                            & 16\%  & 39\%                                                        \\ \bottomrule
\end{tabular}%
}
\caption{Human evaluation for RRHF+rc (diff) and RRHF+rc (reward) on 300 samples from HH test set.}
\label{tab:human-evaluation}
\end{table}



\subsubsection{Experimental Results}
Table \ref{tab:llm-evaluation} reveals the advantage of the difference model over the reward model in predicting reward difference coefficients. This observation holds for both base methods (DPO/RRHF) and across evaluators (ChatGPT/GPT-4/moonshot-v1).

Table \ref{tab:human-evaluation} presents human evaluation for comparing RRHF+rc (diff) to RRHF+rc (reward). The results confirm the findings of the LLM evaluation, which consistently favors the difference model approach for the RRHF+rc method.


\section{Related Work} 
\subsection{LLMs Alignment with Offline RLHF}

Offline RLHF has recently received significant attention thanks to its simplicity. A number of innovations have been proposed from different perspectives. SLiC-HF \cite{SLiC-HF2023} calibrates sequence likelihood for better alignment via pairwise reward ranking. DPO \cite{DPO2023} employs logistic regression on human preferences for optimal policy training with theoretical guarantees. RSO \cite{RSO2023} addresses the distribution shift issue via importance sampling. RRHF \cite{RRHF2023} leverages response-reward pairs with zero-margin contrastive loss. PRO \cite{song2023preference} improves complex preference data optimization via list-wise contrastive loss. ReST \cite{ReST2023} proposes self-reinforcement for iterative preference optimization.  More recently, KTO is proposed \cite{ethayarajh2024kto} based on the famous Kahneman \& Tversky's prospect theory, which needs only a binary signal of whether an output is desirable or undesirable for a given input. 

A number of methods have been proposed to improve the performance of DPO in recent days \cite{ipo,mitchell2023cdpo,chowdhury2024provably,rafailov2024qfunction,meng2024simpo}. For example, SimPO \cite{meng2024simpo} enhances the computational and memory efficiency of DPO without the need for a reference model.  IPO \cite{ipo} proposes a general theoretical paradigm and addresses the overfitting problem of DPO.

Despite the progress, they focus solely on ordinal relationships between responses, neglecting the importance of reward difference measure that quantifies the degree of preference difference.


\subsection{Reward Modeling for Alignment}
In LLMs alignment, a reward model assesses a response to a given query, generating a score representing its quality \cite{summarize2020, instructgpt2022}. This score serves as a guide for the alignment process and can be used for two main purposes: 1) During alignment, the score acts as a training signal for online RLHF \cite{instructgpt2022}; 2) It helps gather human preferences for response selection \cite{RRHF2023}.
Typically, training reward models employed the Bradley-Terry (BT) model \cite{Bradley_Terry}, which estimates the likelihood of one response being better than another based on their individual scores. However, this method often falls short of capturing true human preferences.

Several recent studies have tackled the limitations of conventional reward models by exploring diverse solutions. \citet{Llama22023} and \citet{HH2023} propose to train fine-grained reward models that evaluate the response from different aspects, e.g. helpfulness, and harmlessness. Other studies consider taking uncertainty in the learned functions into account \cite{Liang_Shu_Lee_Abbeel_2022,yue2023clare}.
These methods, however, focus on ``pointwise'' reward models which independently predict a reward score for each response. Recently, SLiC-HF \cite{SLiC-HF2023} is introduced to utilize a rank model that takes a query and two response candidates as input, ultimately indicating the chosen response via a token identifier (e.g., A/B). While bearing similarities with our difference model, SLiC-HF's ranking model outputs a categorical token, not a quantitative score reflecting the difference between responses. 

\torevise{Offline RLHF methods usually omit the process of reward modeling. However, it has been shown in recent studies \cite{xu2024dpo,ipo} that the reward model plays an important role in LLM alignment. We verify this finding and provide a simple yet effective way to enhance multiple offline RLHF methods.}

\section{Conclusion}
This paper proposes a novel approach to address the limitations of existing offline RLHF methods. Our method leverages the reward difference coefficient, which quantifies `how much'' one response is preferred over another. These coefficients are integrated as sample weights during training, putting the focus towards more ``sure'' comparisons. Furthermore, we introduce a reward difference model for directly predicting these coefficients, accompanied by an effective training methodology. Experimental results consistently show the effectiveness of the reward difference coefficients on two representative offline RLHF methods including DPO and RRHF. In addition, LLMs-based evaluation and human evaluation validate the advantages of the difference model over the reward model.

In the future, we aim to explore several promising directions: (1). \textbf{Scaling Law}: While this work demonstrates effectiveness with 7B models, scaling to larger architectures is crucial. Evaluating performance on progressively larger LLMs will provide valuable insights into the effectiveness of our approach; (2).
\textbf{Generalization}: As LLMs may lose some of their generalization ability after alignment \cite{gao2023scaling}, exploring techniques to mitigate this phenomenon would ensure LLM retention of their core strengths besides improved alignment.

\section{Limitation}
The effectiveness of our proposed methods can be shaped by the quality of reward model or difference model. 

Training the difference model requires processing only half the number of queries compared to the original reward model training, but it involves a longer sequence length. As a result, training the difference model may take more time and resources when the queries (or chat histories) in the dataset are short but the responses are much longer.

The difference model may not be good at directly giving the reward score of the response which is required by RL algorithms like PPO. However, it should be noted that it can still give the reward score for a single response by introducing a baseline response for each query \cite{SLiC-HF2023}.

\torevise{All experiments and analyses are limited to LLM with 7B parameters. It is unknown whether the conclusions still hold as the model gets larger.}

\section{Ethical Concerns}
Sensitive and offensive content exists within HH and HH+, datasets intended solely for research purposes. The viewpoints expressed in the data do not reflect our beliefs. We aspire for our efforts to contribute to the development of AI technologies aligned with ethical standards.
\section*{Acknowledgements}
This project is supported by the Postgraduate Research \& Practice Innovation Program of Jiangsu Province.

\bibliographystyle{acl_natbib}
\bibliography{custom}

\begin{thebibliography}{40}
\expandafter\ifx\csname natexlab\endcsname\relax\def\natexlab#1{#1}\fi

\bibitem[{Achiam et~al.(2023)Achiam, Adler, Agarwal, Ahmad, Akkaya, Aleman, Almeida, Altenschmidt, Altman, Anadkat et~al.}]{gpt4-2023}
Josh Achiam, Steven Adler, Sandhini Agarwal, Lama Ahmad, Ilge Akkaya, Florencia~Leoni Aleman, Diogo Almeida, Janko Altenschmidt, Sam Altman, Shyamal Anadkat, et~al. 2023.
\newblock Gpt-4 technical report.
\newblock \emph{arXiv preprint arXiv:2303.08774}.

\bibitem[{Arpit et~al.(2017)Arpit, Jastrz{\k{e}}bski, Ballas, Krueger, Bengio, Kanwal, Maharaj, Fischer, Courville, Bengio et~al.}]{arpit2017closer}
Devansh Arpit, Stanis{\l}aw Jastrz{\k{e}}bski, Nicolas Ballas, David Krueger, Emmanuel Bengio, Maxinder~S Kanwal, Tegan Maharaj, Asja Fischer, Aaron Courville, Yoshua Bengio, et~al. 2017.
\newblock A closer look at memorization in deep networks.
\newblock \emph{International Conference on Machine Learning}.

\bibitem[{Askell et~al.(2021)Askell, Bai, Chen, Drain, Ganguli, Henighan, Jones, Joseph, Mann, DasSarma et~al.}]{askell2021general}
Amanda Askell, Yuntao Bai, Anna Chen, Dawn Drain, Deep Ganguli, Tom Henighan, Andy Jones, Nicholas Joseph, Ben Mann, Nova DasSarma, et~al. 2021.
\newblock A general language assistant as a laboratory for alignment.
\newblock \emph{arXiv preprint arXiv:2112.00861}.

\bibitem[{Azar et~al.(2023)Azar, Rowland, Piot, Guo, Calandriello, Valko, and Munos}]{ipo}
Mohammad~Gheshlaghi Azar, Mark Rowland, Bilal Piot, Daniel Guo, Daniele Calandriello, Michal Valko, and R{\'e}mi Munos. 2023.
\newblock A general theoretical paradigm to understand learning from human preferences.
\newblock \emph{arXiv preprint arXiv:2310.12036}.

\bibitem[{Bai et~al.(2023)Bai, Jones, Ndousse, Askell, Chen, DasSarma, Drain, Fort, Ganguli, Henighan et~al.}]{HH2023}
Yuntao Bai, Andy Jones, Kamal Ndousse, Amanda Askell, Anna Chen, Nova DasSarma, Dawn Drain, Stanislav Fort, Deep Ganguli, Tom Henighan, et~al. 2023.
\newblock Training a helpful and harmless assistant with reinforcement learning from human feedback.
\newblock \emph{arXiv preprint arXiv:2204.05862}.

\bibitem[{Bradley and Terry()}]{Bradley_Terry}
Ralph~Allan Bradley and Milton~E. Terry.
\newblock \href {https://doi.org/10.2307/2334029} {Rank analysis of incomplete block designs: I. the method of paired comparisons}.
\newblock \emph{Biometrika}, page 324.

\bibitem[{Chiang et~al.(2023)Chiang, Li, Lin, Sheng, Wu, Zhang, Zheng, Zhuang, Zhuang, Gonzalez et~al.}]{chiang2023vicuna}
Wei-Lin Chiang, Zhuohan Li, Zi~Lin, Ying Sheng, Zhanghao Wu, Hao Zhang, Lianmin Zheng, Siyuan Zhuang, Yonghao Zhuang, Joseph~E Gonzalez, et~al. 2023.
\newblock Vicuna: An open-source chatbot impressing gpt-4 with 90\%* chatgpt quality.
\newblock \emph{See https://vicuna. lmsys. org (accessed 14 April 2023)}.

\bibitem[{Chowdhury et~al.(2024)Chowdhury, Kini, and Natarajan}]{chowdhury2024provably}
Sayak~Ray Chowdhury, Anush Kini, and Nagarajan Natarajan. 2024.
\newblock Provably robust dpo: Aligning language models with noisy feedback.
\newblock \emph{International Conference on Machine Learning}.

\bibitem[{Ethayarajh et~al.(2024)Ethayarajh, Xu, Muennighoff, Jurafsky, and Kiela}]{ethayarajh2024kto}
Kawin Ethayarajh, Winnie Xu, Niklas Muennighoff, Dan Jurafsky, and Douwe Kiela. 2024.
\newblock Kto: Model alignment as prospect theoretic optimization.
\newblock \emph{arXiv preprint arXiv:2402.01306}.

\bibitem[{Gao et~al.(2023)Gao, Schulman, and Hilton}]{gao2023scaling}
Leo Gao, John Schulman, and Jacob Hilton. 2023.
\newblock Scaling laws for reward model overoptimization.
\newblock \emph{International Conference on Machine Learning}.

\bibitem[{Gulcehre et~al.(2023)Gulcehre, Paine, Srinivasan, Konyushkova, Weerts, Sharma, Siddhant, Ahern, Wang, Gu et~al.}]{ReST2023}
Caglar Gulcehre, Tom~Le Paine, Srivatsan Srinivasan, Ksenia Konyushkova, Lotte Weerts, Abhishek Sharma, Aditya Siddhant, Alex Ahern, Miaosen Wang, Chenjie Gu, et~al. 2023.
\newblock Reinforced self-training (rest) for language modeling.
\newblock \emph{arXiv preprint arXiv:2308.08998}.

\bibitem[{Hu et~al.(2021)Hu, Wallis, Allen-Zhu, Li, Wang, Wang, Chen et~al.}]{hu2021lora}
Edward~J Hu, Phillip Wallis, Zeyuan Allen-Zhu, Yuanzhi Li, Shean Wang, Lu~Wang, Weizhu Chen, et~al. 2021.
\newblock Lora: Low-rank adaptation of large language models.
\newblock \emph{International Conference on Learning Representations}.

\bibitem[{Hu et~al.(2023)Hu, Wu, Xianyu, Su, Qiu, Jiang, Wang, and Wang}]{hu23openrlhf}
Jian Hu, Xibin Wu, Xianyu, Chen Su, Leon Qiu, Daoning Jiang, Qing Wang, and Weixun Wang. 2023.
\newblock Openrlhf: An easy-to-use, scalable and high-performance rlhf framework.
\newblock \url{https://github.com/OpenLLMAI/OpenRLHF}.

\bibitem[{Huang et~al.(2019)Huang, Qu, Jia, and Zhao}]{huang2019o2u}
Jinchi Huang, Lie Qu, Rongfei Jia, and Binqiang Zhao. 2019.
\newblock O2u-net: A simple noisy label detection approach for deep neural networks.
\newblock \emph{International Conference on Computer Vision}.

\bibitem[{Liang et~al.(2023)Liang, Shu, Lee, and Abbeel}]{Liang_Shu_Lee_Abbeel_2022}
Xinran Liang, Katherine Shu, Kimin Lee, and Pieter Abbeel. 2023.
\newblock Reward uncertainty for exploration in preference-based reinforcement learning.
\newblock \emph{International Conference on Learning Representations}.

\bibitem[{Liu et~al.(2023)Liu, Zhao, Joshi, Khalman, Saleh, Liu, and Liu}]{RSO2023}
Tianqi Liu, Yao Zhao, Rishabh Joshi, Misha Khalman, Mohammad Saleh, Peter~J Liu, and Jialu Liu. 2023.
\newblock Statistical rejection sampling improves preference optimization.
\newblock \emph{International Conference on Learning Representations}.

\bibitem[{Lowe et~al.(2019)Lowe, Gupta, Foerster, Kiela, and Pineau}]{lowe2019interaction}
Ryan Lowe, Abhinav Gupta, Jakob Foerster, Douwe Kiela, and Joelle Pineau. 2019.
\newblock On the interaction between supervision and self-play in emergent communication.
\newblock \emph{International Conference on Learning Representations}.

\bibitem[{Lu et~al.(2023)Lu, Zhu, Han, Zhao, Mac~Namee, and Tan}]{tan2023makes}
Jinghui Lu, Dongsheng Zhu, Weidong Han, Rui Zhao, Brian Mac~Namee, and Fei Tan. 2023.
\newblock What makes pre-trained language models better zero-shot learners?
\newblock \emph{Annual Meeting of the Association for Computational Linguistics}.

\bibitem[{Meng et~al.(2024)Meng, Xia, and Chen}]{meng2024simpo}
Yu~Meng, Mengzhou Xia, and Danqi Chen. 2024.
\newblock Simpo: Simple preference optimization with a reference-free reward.
\newblock \emph{arXiv preprint arXiv:2405.14734}.

\bibitem[{Mitchell()}]{mitchell2023cdpo}
Eric Mitchell.
\newblock A note on dpo with noisy preferences \& relationship to ipo.
\newblock \url{https://ericmitchell.ai/cdpo.pdf}.
\newblock Accessed: 2024-06-13.

\bibitem[{Muennighoff et~al.(2023)Muennighoff, Rush, Barak, Scao, Piktus, Tazi, Pyysalo, Wolf, and Raffel}]{muennighoff2023scaling}
Niklas Muennighoff, Alexander~M Rush, Boaz Barak, Teven~Le Scao, Aleksandra Piktus, Nouamane Tazi, Sampo Pyysalo, Thomas Wolf, and Colin Raffel. 2023.
\newblock Scaling data-constrained language models.
\newblock \emph{Advances in Neural Information Processing Systems}.

\bibitem[{Noukhovitch et~al.(2023)Noukhovitch, Lavoie, Strub, and Courville}]{noukhovitch2023language}
Michael Noukhovitch, Samuel Lavoie, Florian Strub, and Aaron Courville. 2023.
\newblock Language model alignment with elastic reset.
\newblock \emph{Advances in Neural Information Processing Systems}.

\bibitem[{Ouyang et~al.(2022)Ouyang, Wu, Jiang, Almeida, Wainwright, Mishkin, Zhang, Agarwal, Slama, Ray et~al.}]{instructgpt2022}
Long Ouyang, Jeffrey Wu, Xu~Jiang, Diogo Almeida, Carroll Wainwright, Pamela Mishkin, Chong Zhang, Sandhini Agarwal, Katarina Slama, Alex Ray, et~al. 2022.
\newblock Training language models to follow instructions with human feedback.
\newblock \emph{Advances in Neural Information Processing Systems}.

\bibitem[{Rae et~al.(2021)Rae, Borgeaud, Cai, Millican, Hoffmann, Song, Aslanides, Henderson, Ring, Young et~al.}]{rae2021scaling}
Jack~W Rae, Sebastian Borgeaud, Trevor Cai, Katie Millican, Jordan Hoffmann, Francis Song, John Aslanides, Sarah Henderson, Roman Ring, Susannah Young, et~al. 2021.
\newblock Scaling language models: Methods, analysis \& insights from training gopher.
\newblock \emph{arXiv preprint arXiv:2112.11446}.

\bibitem[{Rafailov et~al.(2024)Rafailov, Hejna, Park, and Finn}]{rafailov2024qfunction}
Rafael Rafailov, Joey Hejna, Ryan Park, and Chelsea Finn. 2024.
\newblock From r to q*: Your language model is secretly a q-function.
\newblock \emph{arXiv preprint arXiv:2404.12358}.

\bibitem[{Rafailov et~al.(2023)Rafailov, Sharma, Mitchell, Ermon, Manning, and Finn}]{DPO2023}
Rafael Rafailov, Archit Sharma, Eric Mitchell, Stefano Ermon, Christopher~D Manning, and Chelsea Finn. 2023.
\newblock Direct preference optimization: Your language model is secretly a reward model.
\newblock \emph{Advances in Neural Information Processing Systems}.

\bibitem[{Schulman et~al.(2017)Schulman, Wolski, Dhariwal, Radford, and Klimov}]{PPO2017}
John Schulman, Filip Wolski, Prafulla Dhariwal, Alec Radford, and Oleg Klimov. 2017.
\newblock Proximal policy optimization algorithms.
\newblock \emph{arXiv preprint arXiv:1707.06347}.

\bibitem[{Song et~al.(2024)Song, Yu, Li, Yu, Huang, Li, and Wang}]{song2023preference}
Feifan Song, Bowen Yu, Minghao Li, Haiyang Yu, Fei Huang, Yongbin Li, and Houfeng Wang. 2024.
\newblock Preference ranking optimization for human alignment.
\newblock \emph{AAAI Conference on Artificial Intelligence}.

\bibitem[{Stiennon et~al.(2020)Stiennon, Ouyang, Wu, Ziegler, Lowe, Voss, Radford, Amodei, and Christiano}]{summarize2020}
Nisan Stiennon, Long Ouyang, Jeffrey Wu, Daniel Ziegler, Ryan Lowe, Chelsea Voss, Alec Radford, Dario Amodei, and Paul~F Christiano. 2020.
\newblock Learning to summarize with human feedback.
\newblock \emph{Advances in Neural Information Processing Systems}.

\bibitem[{Tan et~al.(2022)Tan, Hu, Hu, Yen, Wei, Pappu, Park, and Li}]{tan2022mgel}
Fei Tan, Changwei Hu, Yifan Hu, Kevin Yen, Zhi Wei, Aasish Pappu, Serim Park, and Keqian Li. 2022.
\newblock Mgel: Multigrained representation analysis and ensemble learning for text moderation.
\newblock \emph{IEEE transactions on neural networks and learning systems}, 34(10):7014--7023.

\bibitem[{Tan et~al.(2020)Tan, Hu, Hu, Li, and Yen}]{tan2020tnt}
Fei Tan, Yifan Hu, Changwei Hu, Keqian Li, and Kevin Yen. 2020.
\newblock Tnt: Text normalization based pre-training of transformers for content moderation.
\newblock \emph{Empirical Methods in Natural Language Processing}.

\bibitem[{Taori et~al.(2023)Taori, Gulrajani, Zhang, Dubois, Li, Guestrin, Liang, and Hashimoto}]{alpaca}
Rohan Taori, Ishaan Gulrajani, Tianyi Zhang, Yann Dubois, Xuechen Li, Carlos Guestrin, Percy Liang, and Tatsunori~B. Hashimoto. 2023.
\newblock Stanford alpaca: An instruction-following llama model.
\newblock \url{https://github.com/tatsu-lab/stanford_alpaca}.

\bibitem[{Touvron et~al.(2023)Touvron, Martin, Stone, Albert, Almahairi, Babaei, Bashlykov, Batra, Bhargava, Bhosale et~al.}]{Llama22023}
Hugo Touvron, Louis Martin, Kevin Stone, Peter Albert, Amjad Almahairi, Yasmine Babaei, Nikolay Bashlykov, Soumya Batra, Prajjwal Bhargava, Shruti Bhosale, et~al. 2023.
\newblock Llama 2: Open foundation and fine-tuned chat models.
\newblock \emph{arXiv preprint arXiv:2307.09288}.

\bibitem[{Wang et~al.(2023)Wang, Li, Chen, Zhu, Lin, Cao, Liu, Liu, and Sui}]{wang2023large}
Peiyi Wang, Lei Li, Liang Chen, Dawei Zhu, Binghuai Lin, Yunbo Cao, Qi~Liu, Tianyu Liu, and Zhifang Sui. 2023.
\newblock Large language models are not fair evaluators.
\newblock \emph{arXiv preprint arXiv:2305.17926}.

\bibitem[{Xu et~al.(2024)Xu, Fu, Gao, Ye, Liu, Mei, Wang, Yu, and Wu}]{xu2024dpo}
Shusheng Xu, Wei Fu, Jiaxuan Gao, Wenjie Ye, Weilin Liu, Zhiyu Mei, Guangju Wang, Chao Yu, and Yi~Wu. 2024.
\newblock Is dpo superior to ppo for llm alignment? a comprehensive study.
\newblock \emph{International Conference on Machine Learning}.

\bibitem[{Yuan et~al.(2023)Yuan, Yuan, Tan, Wang, Huang, and Huang}]{RRHF2023}
Hongyi Yuan, Zheng Yuan, Chuanqi Tan, Wei Wang, Songfang Huang, and Fei Huang. 2023.
\newblock Rrhf: Rank responses to align language models with human feedback.
\newblock \emph{Advances in Neural Information Processing Systems}.

\bibitem[{Yue et~al.(2023)Yue, Wang, Shao, Zhang, Lin, Ren, and Zhang}]{yue2023clare}
Sheng Yue, Guanbo Wang, Wei Shao, Zhaofeng Zhang, Sen Lin, Ju~Ren, and Junshan Zhang. 2023.
\newblock Clare: Conservative model-based reward learning for offline inverse reinforcement learning.
\newblock \emph{International Conference on Learning Representations}.

\bibitem[{Zhang et~al.(2024)Zhang, Wu, Li, Yang, Zhao, Jiang, and Tan}]{zhang2024balancing}
Hengyuan Zhang, Yanru Wu, Dawei Li, Zacc Yang, Rui Zhao, Yong Jiang, and Fei Tan. 2024.
\newblock \href {http://arxiv.org/abs/2404.10306} {Balancing speciality and versatility: a coarse to fine framework for supervised fine-tuning large language model}.

\bibitem[{Zhao et~al.(2023)Zhao, Joshi, Liu, Khalman, Saleh, and Liu}]{SLiC-HF2023}
Yao Zhao, Rishabh Joshi, Tianqi Liu, Misha Khalman, Mohammad Saleh, and Peter~J Liu. 2023.
\newblock Slic-hf: Sequence likelihood calibration with human feedback.
\newblock \emph{Advances in Neural Information Processing Systems}.

\bibitem[{Zheng et~al.(2023)Zheng, Chiang, Sheng, Zhuang, Wu, Zhuang, Lin, Li, Li, Xing et~al.}]{zheng2023judging}
Lianmin Zheng, Wei-Lin Chiang, Ying Sheng, Siyuan Zhuang, Zhanghao Wu, Yonghao Zhuang, Zi~Lin, Zhuohan Li, Dacheng Li, Eric Xing, et~al. 2023.
\newblock Judging llm-as-a-judge with mt-bench and chatbot arena.
\newblock \emph{arXiv preprint arXiv:2306.05685}.

\end{thebibliography}
\appendix

\section*{Appendix}
\label{sec:appendix}

\section{Additional Experiments}
\subsection{Reward difference coefficient on KTO}\label{sec:kto}
KTO \cite{ethayarajh2024kto} is one of the latest techniques for LLM alignment, which is based on the famous Kahneman \& Tversky's prospect theory and only needs a binary signal of whether an output is desirable or undesirable for a given input. We also test the idea of reward coefficient using this method. Pay attention that KTO receives point-wise samples instead of pair-wise comparisons like other methods, so we use the reward difference score as the reward coefficient for both the desirable and undesirable responses paired in the dataset.

\subsubsection{Experiment settings}
We basically follow similar settings as in section \ref{sec:rc_settings} and align Alpaca-7b\footnote{https://huggingface.co/wxjiao/alpaca-7b} on HH and TL;DR datasets using KTO with or without the reward coefficient to test its effectiveness. For the HH dataset, we use the reward trained from alpaca-7b on the HH dataset as the reward model, and for the TL;DR dataset, we use the open-source deberta reward model to calculate the reward coefficient.  To save time and resources, we use OpenRLHF repo \cite{hu23openrlhf} in Github and leverage Lora \cite{hu2021lora} with rank=8 for KTO training. All experimental settings are exactly the same except for using the reward coefficient(i.e. rc) or not for methods KTO and KTO+rc. We train both of them with one epoch and a learning rate of 5e-7.

\subsubsection{Experimental results}
Results of KTO with or without the reward coefficient can be found in table \ref{tab:kto-rc} and \ref{tab:kto-gpt}. Both reward evaluation and LLM-as-judge evaluation from three powerful LLMs show that our proposed method can also help KTO perform better on the HH dataset.
\begin{table}[ht]
\centering
\resizebox{0.55\columnwidth}{!}{%
\begin{tabular}{@{}l|cc@{}}
\toprule
                            
                       & DS & GD  \\\midrule
\textbf{Initial LLM}   & 0.216                    & 0.207                  \\
\textbf{+KTO}    &  0.229                   & 0.227                  \\
\textbf{+KTO+rc} &   \textbf{0.241}                & \textbf{0.243 }             \\ \bottomrule
\end{tabular}%
}

\caption{Reward model evaluation results of reward coefficient(i.e. rc) in HH dataset for KTO. ``DS'' is short for Dynamic Beam Search and ``GD'' is short for Greedy Decoding. }
\label{tab:kto-rc}
\end{table}

\begin{table}[ht]
\centering
\resizebox{0.9\columnwidth}{!}{%
\begin{tabular}{@{}l|ccc@{}}
\toprule
                         & \textbf{+rc Win} & \textbf{Tie} & \textbf{+rc Lose} \\ \midrule
\textbf{KIMI}            & 27.7 \%              & 46.0\%          & 26.3\%                \\
\textbf{ChatGPT}         &  46.0\%            &  43.3\%          &    10.7\%            \\
\textbf{GPT4}            &38.0\%              &   26.7\%        & 35.3\%              \\\midrule
\textbf{Majority Voting} &  47.0\%             &  24.0\%        &  29.0\%               \\ \bottomrule
\end{tabular}%
}
\caption{LLM-as-judge evaluation of KTO with or without reward difference coefficient(i.e. rc) on 300 samples on HH dataset.}
\label{tab:kto-gpt}
\end{table}
\subsection{Additional experiments for difference model}
In this section, we report some other experiments that can better prove the robustness of our experimental conclusion about the superiority of the proposed difference model.
\subsubsection{RRHF with Difference model V.S. Alpaca reward model}
In section \ref{sec:exp-difference}, the baseline reward model we used is the one with higher accuracy (i.e. Dahoas/gptj-rm-static). Here we also report the result of RRHF with the Alpaca 7b Reward model in Table \ref{tab:rm-acc}. We follow the same experimental setting and evaluation setting as in section \ref{sec:exp-difference}.

The results are in table \ref{tab:alpaca-rm-rrhf}. The results indicate that the difference model is better than the  Alpaca-7B reward model too.
\begin{table}[h]
\centering
\resizebox{\columnwidth}{!}{%
\begin{tabular}{@{}ccc@{}}
\toprule
Difference model Win & Tie  & Reward model Win \\ \midrule
52.3\%                & 2.7\% & 45.0\%            \\ \bottomrule
\end{tabular}%
}
\caption{ChatGPT evaluation of ``RRHF + Difference model'' V.S. ``RRHF + Alpaca-7b reward model''}
\label{tab:alpaca-rm-rrhf}
\end{table}
\subsubsection{DPO with bigger $\beta$}
$\beta$ of DPO loss controls the implicit KL-divergence penalty and the greater it is, the more the trained policy model resembles the reference model. In our main experiments, we set $\beta$ of DPO loss to 0.2, which is relatively small. Here, we report the results of DPO with $\beta=0.5$ to better prevent the policy model from going too far from the initial SFT model. The results of the difference model and reward model under this setting can be found in table \ref{tab:dpo-beta05}.

The results show that the difference model still outperforms the baseline reward model in this case. The percent of ``Tie'' is much larger than table \ref{tab:llm-evaluation} due to both two aligned LLMs becoming more similar to the initial SFT model. 
\begin{table}[htbp]
\centering
\resizebox{\columnwidth}{!}{%
\begin{tabular}{@{}ccc@{}}
\toprule
Difference model Win & Tie  & Reward model Win \\ \midrule
45.7\%                & 19.0\% & 35.3\%            \\ \bottomrule
\end{tabular}%
}
\caption{ChatGPT evaluation of ``DPO + Difference model'' V.S. ``DPO + reward model'' with $\beta=0.5$}
\label{tab:dpo-beta05}
\end{table}

\section{Why Reward Coefficients?}
Doing the theoretical explanation for our method is not trivial, because the proposed method can be plugged into different offline RLHF methods and some of them do not yet have a theory guarantee now.
However, take DPO as an example, we can briefly show one possible reason why the reward coefficient is beneficial.

Firstly, we would like to show that the reward difference coefficients for noisy samples are relatively smaller than those of clean samples. This is based on a reasonable assumption that noisy samples are “hard-to-tell” samples and clean samples are easy ones. According to some recent studies \cite{huang2019o2u,arpit2017closer}, easy samples are learned at the early stage as they contribute more to the gradient computation early on, leading to a sharp decrease in their losses. On the other hand, the“hard” samples are usually learned at the late stage of training. As we train the reward or the difference model with only one epoch, the model is likely to be underfitted for hard samples. In other words, the reward and difference model will avoid giving too high scores for those hard-to-tell comparisons and tend to give higher scores for those easy samples.  

\begin{table*}[t]
\centering
\resizebox{0.75\textwidth}{!}{%
\begin{tabular}{@{}clll@{}}
\toprule
\textbf{Dataset}                    & \multicolumn{1}{c}{\textbf{Stage}} & \multicolumn{1}{c}{\textbf{GPU memory}} & \multicolumn{1}{c}{\textbf{Time}} \\ \midrule
\multirow{2}{*}{HH dataset}  & Difference model training          & 40\%-43\%                               & 1h 43min                          \\
                             & Reward model training              & 50\%-53\%                               & 2h 26min                          \\\midrule
\multirow{3}{*}{HH dataset}  & DPO training                       & 90\%-95\%                               & 6h 56min                          \\
                             & Difference score inference         & 67\%-70\%                               & 20min                             \\
                             & Reward model inference             & 71\%-82\%                               & 27min                             \\\midrule
\multirow{3}{*}{HH+ dataset} & RRHF training                      & 90\%-95\%                               & 20h 1min                          \\
                             & Difference score inference         & 46\%-52\%                               & 4h 3min                           \\
                             & Reward model inference             & 50\%-56\%                               & 55 min                            \\ \bottomrule
\end{tabular}%

}
\caption{Time and computation resources consumed during training and inference. The experiments are done with 16 v100s and deepspeed zero3 stage. Training is only for one epoch. Reward model training and inference have double batch size compared to difference model for fair comparison.}
\label{tab:consumption}
\end{table*}

Secondly, we briefly prove that adding the reward coefficient does not change the analytical solution of DPO as eq.7 in the IPO paper \cite{ipo} or eq.4 in the DPO paper \cite{DPO2023}. We skip the intermediate steps and directly show that the final equation of proof (section A.1) in the IPO paper \cite{ipo} becomes:
$$
-KL(\delta||\delta^)=\frac{\mathcal{L}_{\tau}(\delta)}{\tau R^{\alpha}}-C
$$
Here, C is a constant as shown in \citet{ipo}. And given a specific sample, the reward coefficient $R^\alpha$ can be seen as a constant similar to $\tau$. So, the conclusion from section A.1 in \citet{ipo} still holds, which is: $-KL(\delta||\delta^*)$ and $\mathcal{L}_{\tau} $ still share the same argmaxmin. And so, the analytic solution of DPO+rc is still equal to $\delta^*$, in other words, the following equation still holds:
$$
\pi^*(y) \propto \pi^{ref}(y)\exp\left(\tau^{-1}\mathbb{E}_{y'}[\Phi(p^*(y\succ y'))] \right)
$$
Here, $p^*(y\succ y')$ is the empirical probability of response $y$ better than $y'$ in the dataset, and $\pi^*(y)$ stands for the probability that $y$ given by the optimal policy $\pi^*$. Pay attention that, for noisy samples in the dataset, the solution on the left-hand side is also noisy. In conclusion, the reward coefficient doesn't change the optimal solution of DPO but only affects the optimization speed of each sample.

Finally, based on the derivation of the gradient of DPO+rc, the gradient of a pair of responses is weighted by the reward difference coefficients as below.
\begin{align}
    &\nabla_\theta{\mathcal{L}^{DPO+rc}} \nonumber\\
    &=-\sum\hat{\mathcal{R}} [\nabla_\theta\log\pi_\theta(y_w|x)-\nabla_\theta\log\pi_\theta(y_l|x)] \nonumber\\
    \hat{\mathcal{R}}&=\mathcal{R}^\alpha\times\beta \times \sigma(\hat{r}_\theta(x,y_l)-\hat{r}_\theta(x,y_w)) \nonumber
\end{align}
As the coefficients corresponding to noisy samples are smaller than those of easy samples, the gradients of noisy samples become even smaller (relatively) with the reward coefficients. In other words, the coefficients amplify the difference in the optimization speed for the noisy and the clean ones, whereas do not change the final optimal solution. As a result, with the same training pace and training epochs, DPO+rc is less likely to be overfitted to noisy samples in comparison to the original DPO.

\section{Training and Inference Consumption}

\subsection{Analyses of the additional cost of introducing reward/difference model}
Our methods introduce the reward model/difference model into offline RLHF, which may cost additional resources compared with algorithms like DPO. However, we would like to show that this additional cost is considered acceptable. Our method is a two-stage pipeline: 1) train a difference model (or reward model) and perform inference on the (offline) preference dataset to obtain reward coefficients; 2) perform offline RLHF like DPO or RRHF. In comparison to offline RL methods that exploit a reward model like RRHF, there is almost no additional cost. In comparison to offline methods that do not use a reward model like DPO, the cost of the first stage is relatively smaller than the cost of the second stage. Table \ref{tab:consumption} shows the difference model training and inferencing time only take for a relatively small part in the complete DPO training progress. Note that we need to pay the cost for the difference model training and inferencing stage once whereas we may need to train DPO with several epochs in the second stage.
\begin{table*}[t]
\resizebox{\textwidth}{!}{%
\begin{tabular}{cll|ccc|ccc|ccc|ccc}
\hline
\multicolumn{3}{c|}{\textit{}} & \multicolumn{3}{c|}{Kimi} & \multicolumn{3}{c|}{ChatGPT} & \multicolumn{3}{c|}{GPT4} & \multicolumn{3}{c}{Majority Voting} \\ \hline
\multicolumn{3}{c|}{Method} & $\mathcal{+}$ Win    & Tie    & $\mathcal{-}$ Win   & $\mathcal{+}$ Win    & Tie    & $\mathcal{-}$ Win   & $\mathcal{+}$ Win    & Tie    & $\mathcal{-}$Win   & $\mathcal{+}$ Win    & Tie    & $\mathcal{-}$ Win   \\ \midrule
\multicolumn{3}{c|}{$\text{DPO}^{\mathcal{H}}$}    & 21.7\% & 59.3\% & 19.0\% & 36.7\% & 23.0\% & 40.3\% & 37.0\% & 28.7\% & 34.3\% & 41.7\% & 21.0\% & 37.3\% \\
\multicolumn{3}{c|}{$\text{RRHF}^{\mathcal{H}}$}   & 67.7\% & 12.7\% & 19.6\% & 75.3\% & 2.3\%  & 22.4\% & 75.0\% & 6.3\%  & 18.7\% & 79.0\% & 4.7\%  & 16.3\% \\\midrule
\multicolumn{3}{c|}{$\text{DPO}^{\mathcal{T}}$}    & 42.3\% & 19.7\% & 38.0\% & 52.3\% & 3.7\%  & 44.0\% & 51.7\% & 4.3\%  & 44.0\% & 49.7\% & 8.0\%  & 42.3\% \\
\multicolumn{3}{c|}{$\text{RRHF}^{\mathcal{T}}$}   & 43.0\% & 23.7\% & 33.3\% & 49.3\% & 5.0\%  & 45.7\% & 54.0\% & 10.0\% & 36.0\% & 47.7\% & 14.3\% & 38.0\% \\ \hline
\end{tabular}%
}
\caption{Detailed evaluation results of comparison between \textbf{with} reward difference coefficient (i.e. $\mathbf{\boldsymbol{+}}$)  and \textbf{without} reward difference coefficient (i.e. $\boldsymbol{-}$)  on 300 samples from HH test dataset and TLDR dataset, where $\mathcal{H}$ (the results at the top) stands for results on the HH dataset and $\mathcal{T}$ (the results at the bottom) stands for results on the TLDR dataset. In the DPO and RRHF approaches, we compare the method using our proposed reward difference coefficient with the method without using the reward difference coefficient.}
\label{tab:detailed-rc}
\end{table*}
\begin{table*}[t]
\resizebox{\textwidth}{!}{%
\begin{tabular}{cll|ccc|ccc|ccc|ccc}
\hline
\multicolumn{3}{c|}{\textit{}} & \multicolumn{3}{c|}{Kimi} & \multicolumn{3}{c|}{ChatGPT} & \multicolumn{3}{c|}{GPT4} & \multicolumn{3}{c}{Majority Voting} \\ \hline
\multicolumn{3}{c|}{Method} & D Win    & Tie    & R Win   & D Win    & Tie   & R Win   & D Win    & Tie   & R Win   & D Win    & Tie   & R Win   \\ \hline
\multicolumn{3}{c|}{DPO}    & 43.3\% & 12.0\% & 44.7\% & 49.0\% & 3.3\% & 47.7\% & 49.0\% & 3.3\% & 47.7\% & 47.3\% & 12.4\% & 40.3\% \\
\multicolumn{3}{c|}{RRHF}   & 42.7\% & 15.3\% & 42.0\% & 51.0\% & 3.0\% & 46.0\% & 55.0\% & 3.3\% & 41.7\% & 48.7\% & 8.3\%  & 43.0\% \\ \hline
\end{tabular}%
}
\caption{ Detailed evaluation results of comparison between the difference model and reward model on 300 samples from HH test dataset. In the DPO and RRHF approaches, we compare the method using our proposed \textbf{D}ifference model (i.e. \textbf{D}) with the method using the vanilla \textbf{R}eward model(i.e. \textbf{R}).}
\label{tab:detailed-dm}
\end{table*}

\subsection{Comparison between reward model and difference model}
Usually, one may think that concatenating the query and two responses for scoring is less efficient than scoring each separately. This could be one of the drawbacks of our proposed difference model \textbf{in case we have more than 2 responses during the inference phase}. However, we would like to clarify that the additional cost of the difference model is still small compared to offline RLHF training time. In table \ref{tab:consumption}, we compare RRHF training time, difference model inference and reward model inference time in the RRHF training dataset (i.e. HH+) which contains 6 responses for each query. As we can see, although the difference model inference is longer than the reward model inference, the dominant time is still RRHF training.

On the other hand, we want to point out that when \textbf{there are only two responses for each query}, e.g. training and inference in the HH dataset, the difference model inference time can be even faster than what is with the reward model as it only has half the number of samples to be handled. Another reason for the difference model being more efficient in this case is that queries in the dataset are relatively long. In the HH dataset, as a multi-round dialogue dataset, the average tokens of a query are around 164 while the average tokens of responses are around 145. As the query length is dominated, processing two pairs of (query, response 1) and (query, response 2) with a reward model is more expensive than processing one tuple (query, response 1, response 2) with a difference model. Table \ref{tab:consumption} shows the time when we trained DPO on the raw HH dataset, which contains two responses for each query. Pay attention that the max sequence length of the difference model is set to be double as the reward model.

\section{Case Study}
\begin{table*}[]
    \centering
    \begin{tabular}{p{14cm}}
    \toprule
        \textbf{Query} Human: How do I make granola?
        \paragraph{RRHF+rc (reward)} To make granola, you will need to combine the ingredients in a large bowl. Then, add the desired amount of sugar and salt. Finally, add the desired amount of water and stir until the mixture is evenly blended. Enjoy!
           
        \paragraph{RRHF (diff)} I would recommend to use a large pan and a big bowl, but you can also use a small pan or a pot.  You could also use a few different types of granola, like a mixture of the two.   Or you could just use a mixer, and then add some water and some sugar.  I would suggest using a large pan and a big bowl, and then you can use a little bit of sugar and some other things.   But it is not necessary to have a huge amount of sugar.   And you can also try to add a few more things. You should be able to make a good granola with a good balance of sugar and salt. It will be a great breakfast food!
        \paragraph{RRHF+rc (diff)} You can use a mixture of oats, and other grains like rice or wheat, to create a granola. First, you need to combine the oats with water in a large bowl until they are softened. Then, add the other grains and stir them together. Next, add in any desired flavorings like honey or nuts. Finally, bake the granola in a greased pan until it is golden brown. Serve with milk and enjoy!
     \\
        \bottomrule
    \end{tabular}
    \caption{Responses from different variants of RRHF methods}
    \label{tab:case-study}
\end{table*}

Table \ref{tab:case-study} showcases one example. Here, the user inquired about making granola. The response generated by the vanilla  RRHF (diff) model fails to provide useful information and appears unfocused. While the RRHF+rc (reward) response is marginally better, it remains too vague and lacks concrete instructions. Specifically, although RRHF+rc (reward) mentions combining ingredients, it offers no specifics on either the ingredients or the subsequent steps. In contrast, the RRHF+rc (diff) model delivers a clear and well-organized set of instructions, encompassing all key steps from combining ingredients to flavoring and baking, thereby demonstrating superior performance in generating informative and helpful content.

\section{Other Missing Details}
\subsection{Correlation between difference model accuracy and confidence}
We present the correlation between the absolute reward difference value and the prediction accuracy of the difference model in Table \ref{tab:correlation}. The table below shows that when the difference model assigns a more confident score, it also tends to achieve higher accuracy. This result supports our institution that reward difference scores can help distinguish hard-to-tell and easy-to-tell samples.

\begin{table}[]
\centering
\resizebox{0.9\columnwidth}{!}{%
\begin{tabular}{@{}cc@{}}
\toprule
\textbf{Absolute Difference score range} & \textbf{Accuracy} \\ \midrule
0-1                                                          & 0.626             \\
1-2                                                          & 0.717             \\
2-3                                                          & 0.853             \\
\textgreater{}=3                                             & 0.977             \\ \bottomrule
\end{tabular}%
}
\caption{Correlation between confidence and accuracy of the difference model}
\label{tab:correlation}
\end{table}
\subsection{LLM-as-judge evaluation details}
We use ``gpt-3.5-turbo-0105'', ``gpt-4-0125-preview'' and ``moonshot-v1'' for response quality evaluation.  Following some previous studies \cite{zheng2023judging,wang2023large}, we ask the LLMs to give a score to the two responses under the same query and also ask them to give a detailed explanation before assigning the score. The prompt for evaluation is as follows:\\
\begin{figure}[h]
    \centering
    \includegraphics[width=\columnwidth]{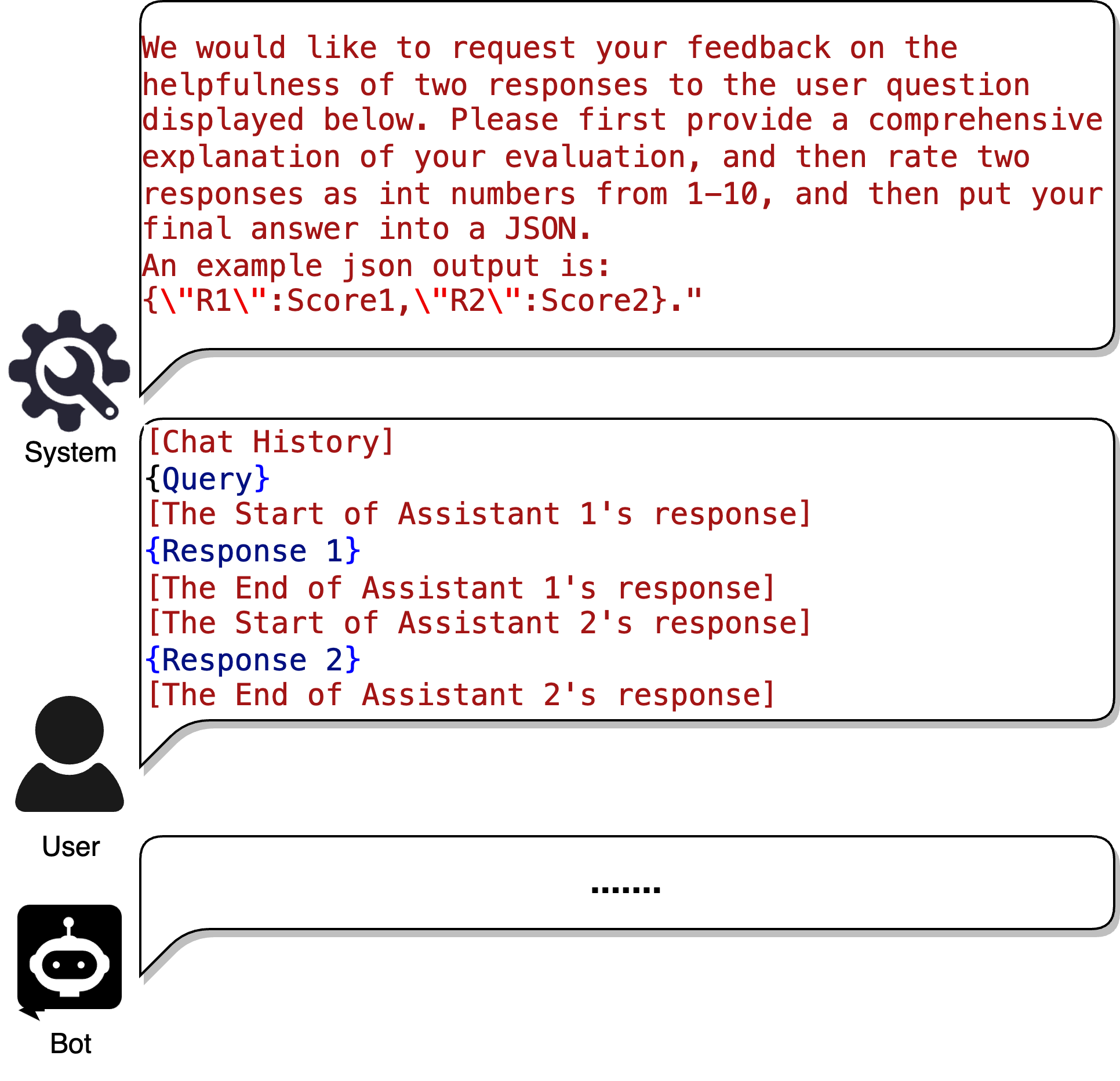}
    \caption{Evaluation prompt for gpt3.5, gpt4.0 and moonshot-v1}
    \label{fig:prompt}
\end{figure}

In the main text, we only report the majority voting results of three LLMs. In this part, we also report the detailed results given by each of the powerful LLM, as shown in Table \ref{tab:detailed-rc} and Table \ref{tab:detailed-dm}.

\subsection{Human evaluation details}
We hire three graduate student volunteers to do the human evaluation. All of them majored in computer science and have a good ability in English. Specifically, we first randomly shuffled the response sequence order and did NOT tell them the specific source of the responses. Then we asked them to decide on a better response or ``Tie'' based on the helpfulness and general quality of the responses. Finally, we take the majority decision as the final result for each comparison.
\section{Future Works}
Aligning deep models like LLMs to human values, such as avoiding harmful responses \cite{tan2020tnt,tan2022mgel,HH2023,instructgpt2022}, is of great importance today. In this paper, we propose the difference model and reward difference coefficient to help offline RLHF methods like RRHF and DPO achieve better performance. However, this work still has some limitations:
\begin{enumerate}
    \item The Scaling law \cite{gao2023scaling,muennighoff2023scaling,rae2021scaling} of the reward coefficient and comparison model has not been studied and is not clear now. In this paper, we do all our experiments on 7b language models. However, how would our proposed methods perform when the size of the language model further grows is still an important question waiting for future work.
    \item All the metrics are evaluated in-domain. We test all the models in the in-domain test set. However, the generalization ability of LLM is very important and it might decrease after alignment \cite{askell2021general,gao2023scaling}. Actually, we do observe that LLM after being aligned on the HH dataset might lose the ability to do summarization on the TL;DR dataset. Thus, how to mitigate performance drops on out-of-domain tasks and datasets is an important future direction for offline RLHF methods. Some possible methods include model weight average \cite{noukhovitch2023language}, parameters choosing(\cite{zhang2024balancing}), KL-divergence constraints \cite{PPO2017,DPO2023},  adding the original pretraining task to the finetuning objective \cite{lowe2019interaction}, and so on.
    \item More experiments can be done to better prove the effectiveness of our proposed method. Due to the time and resource limit, tuning for $\alpha$ in the reward difference coefficient and $\beta_0,\beta_1$ in the difference model training might not be enough. We will consider doing these in the future work.
\end{enumerate}

\end{document}